\documentclass[10pt,twocolumn,letterpaper]{article}

 \usepackage[pagenumbers]{cvpr} % To force page numbers, e.g. for an arXiv version

\usepackage[dvipsnames]{xcolor}

\definecolor{cvprblue}{rgb}{0.21,0.49,0.74}
\usepackage[pagebackref,breaklinks,colorlinks,citecolor=cvprblue]{hyperref}

\def\paperID{EgoVis} % *** Enter the Paper ID here
\def\confName{CVPR}
\def\confYear{2024}

\title{Identification of Conversation Partners from Egocentric Video}

\author{Tobias Dorszewski\\
Technical University of Denmark\\
{\tt\small tobdor@dtu.dk}
\and
 S\o ren A. Fuglsang\\
Copenhagen University Hospital	\\
{\tt\small sorenaf@drcmr.dk}
\and
 Jens Hjortkj\ae r\\
Technical University of Denmark\\
{\tt\small jhjort@dtu.dk}
}

\begin{document}
\maketitle
\begin{abstract}
Communicating in noisy, multi-talker environments is challenging, especially for people with hearing impairments. Egocentric video data can potentially be used to identify a user's conversation partners, which could be used to inform selective acoustic amplification of relevant speakers. Recent introduction of datasets and tasks in computer vision enable progress towards analyzing social interactions from an egocentric perspective. Building on this, we focus on the task of identifying conversation partners from egocentric video and describe a suitable dataset. Our dataset comprises 69 hours of egocentric video of diverse multi-conversation scenarios where each individual was assigned one or more conversation partners, providing the labels for our computer vision task.
This dataset enables the development and assessment of algorithms for identifying conversation partners and evaluating related approaches. Here, we describe the dataset alongside initial baseline results of this ongoing work, aiming to contribute to the exciting advancements in egocentric video analysis for social settings.
\end{abstract}    
\section{Introduction}
\label{sec:intro}

Noisy situations with many people talking simultaneously are very common in everyday life but engaging in conversations in such situations can be a significant challenge. This is particularly the case for people with hearing impairments which can lead to social isolation \cite{Shukla2020HearingReview}.
Accordingly, there is a high incentive to develop technologies that could help those affected. Speech audio separation technologies are rapidly progressing and can offer high-quality acoustic separation and selective amplification of individual speech sources.Yet, a key challenge lies in identifying who the conversation partners are based on wearable sensors. 

%Attempts to solve this problem have included audio-based approaches, electrophysiological sensors and egocentric video.
Using egocentric video is a particularly interesting approach to solving this problem because of the richness of information about the conversation context that can be potentially harnessed. Information about looking directions, mouth movements, postures and interactions between people in the egocentric video could be used to solve the task of identifying social partners. However, it is not clear what the optimal way of integrating this rich information would be. Processing of the egocentric video could be effectively done by employing deep neural networks to integrate available information without the need for explicit definition of relevant features. However, this can potentially limit model interpretability and lead to high computational cost, which is not desirable for a wearable device. Additionally, algorithms trained to identify conversation partners may be prone to overfitting to certain conversation scenarios they were trained on. For instance, egocentric video containing only people in the same conversation group may be much more abundant in datasets, and algorithms may fail to generalize to those competing speaker situations where the technology is most needed.  

In recent years, promising work has introduced new datasets and concepts that enable the analysis of social situations with egocentric video \cite{ Donley2021EasyCom:Environments,Grauman2022Ego4D:Video, Ryan2023EgocentricConversations, JiaThePerspective}. For example, as part of the Ego4D project \cite{Grauman2022Ego4D:Video} the tasks of identifying who is looking at, or talking to the camera wearer were proposed.
This development can yield insights into the challenges posed by complex conversation scenarios with perspectives for smart hearing instrument technology. However, it is not trivial what the best strategy for achieving this goal is. The target of auditory attention may not necessarily be fixed on a single speaker, but could be a group of people. Rather than focusing on a single target at the time, a more distributed approach may be a better amplification strategy for a hypothetical smart hearing device. Based on these considerations we introduce the task of identifying a camera wearer's conversation partners. We define conversation partners as everyone that is part of the camera wearer's conversational group. 

With this work we hope to contribute to the discussion on and development of egocentric computer vision methods for social interactions. Our contributions are:
\begin{itemize}
    \item We introduce the task of identifying conversation partners in the camera wearer's egocentric video
    \item We describe an annotated dataset with diverse multi-conversation scenarios for this task
    \item We present our initial baseline results where we evaluate how accurately conversation partners can be identified based on rudimentary visual features.
\end{itemize}
%------------------------------------------------------------------------
\section{Related Works}
\label{sec:related}
%Analyzing social interactions using computer vision has aimed to develop tools for conversation analysis and support for individuals facing social challenges. Computer Vision tasks in this domain can take the form of face/person detection, active speaker detection/localization or audio-visual speech transcription.
Investigating social interactions from an egocentric video perspective has been the focus for an increasing amount of datasets and tasks.
%The EgoCom dataset \cite{Northcutt2020EgoCom:Dataset} contains 38.5 hours of conversations of 3 individuals and provides labels for speech activity / turn taking and speech transcription.
For example, the EasyCom dataset \cite{Donley2021EasyCom:Environments} features 5h of conversations of 3-5 participants in a noisy environment with a range of labels and the goal of developing solutions to improve the audibility of partners for the camera wearer. Ego4D \cite{Grauman2022Ego4D:Video} includes 45 h of egocentric video during social interaction together with labels for the social interaction tasks “Looking at me” (LAM) and “Talking to me” (TTM). Ryan et al. \cite{Ryan2023EgocentricConversations} used a 20 h dataset and the computer vision task of selective auditory attention localization” (SAAL). On the same dataset, Jia et al. \cite{JiaThePerspective} proposed the task of estimating not just who the camera wearer is listening or speaking to but also who everyone else is listening or speaking to (i.e. estimating the 'exocentric' social graph) based on one egocentric video stream.

All these approaches are highly relevant and complimentary to the task of identifying conversation partners. The main distinction lies in the temporal dimension: conversational groups typically remain stable over longer periods (e.g. minutes), while speech activity and gaze behavior can change rapidly (e.g. seconds). Short-term features based on models tackling the TTM or LAM tasks could provide valuable insights into identifying conversation partners even at time points when they are not talking or looking at the user. Conversely, long-term information on past conversation partners could enhance the performance of TTM or LAM models in multi-conversation settings.

The SAAL task \cite{Ryan2023EgocentricConversations} is closely related to the task of identifying conversation partners. For SAAL the targets of attention are defined as persons “if they are both speaking and in the wearer’s conversation group”\cite{Ryan2023EgocentricConversations}. In contrast, our task focuses on identifying all individuals within the camera wearer’s conversation group and is thus not reliant on active speaker localization (ASL). Ryan et al. also demonstrate that by decoupling SAAL from ASL, their model can learn who is part of the camera wearer’s group \cite{Ryan2023EgocentricConversations}. However, this promising idea is only evaluated in terms of SAAL with perfect ASL and was not pursued further. Our approach separates the long-term task of identifying conversation partners from the short-term task of ASL. Therefore, we expect that the classification of conversation partners may be more stable over time while also not having to make the assumption that auditory attention is only directed towards a speaking group member. This could make the analysis of egocentric video in multi-conversation scenarios more flexible and advantageous in downstream tasks

Most existing datasets feature only one conversation group \cite{ Donley2021EasyCom:Environments} or lack annotations for conversation groups \cite{Grauman2022Ego4D:Video}. This may limit their ability to capture the complexities of multi-conversation situations and may be suboptimal for the task of identifying conversation partners. The SAAL dataset stands out as the only dataset explicitly containing social interactions in more than one simultaneous conversation group, featuring five individuals conversing in two groups \cite{Ryan2023EgocentricConversations}. Yet, a more diverse set of communication contexts is valuable for effectively developing and evaluating a system to identify conversational partners that generalize well across contexts.

\begin{figure}[t]
    % at the top right of page 2
  \centering
    \begin{subfigure}{.8\linewidth}
    \includegraphics[width=\linewidth]{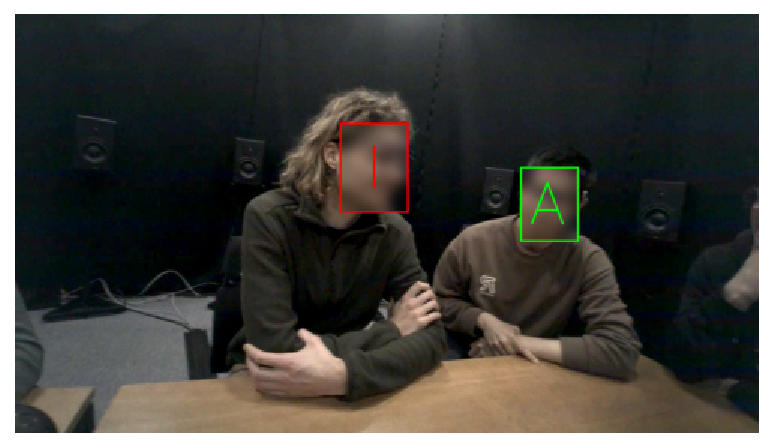}
    \end{subfigure}
        \begin{subfigure}{.6\linewidth}
    \includegraphics[width=\linewidth]{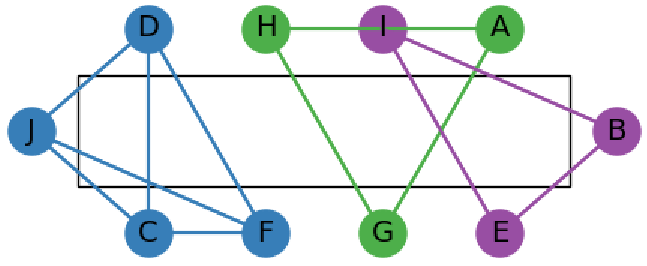}
    \end{subfigure}
   \caption{Example of a frame of egocentric video from our dataset and the group arrangement used in this conversation. The video frame is seen from the perspective of participant G, who is in a conversation with A (green bounding box in the frame) and H (not in view). Approximately 50\% of the dataset feature conversations with some spatial overlap such as the two conversation groups shown on the right in the seating plan.}
   \label{fig:frame}
\end{figure}
\section{The Dataset}
\label{sec:dataset}
\begin{figure*}[ht]
% should be at the top of page 3
  \centering
  \begin{subfigure}{0.31\linewidth}
    \includegraphics[width=\linewidth]{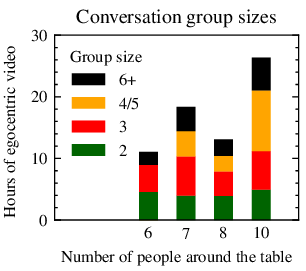}
  \end{subfigure}
  \hfill
    \begin{subfigure}{0.31\linewidth}
    \includegraphics[width=\linewidth]{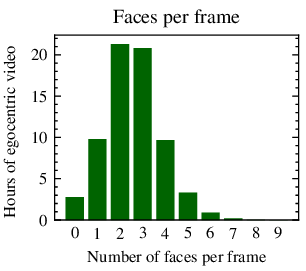}
  \end{subfigure}
  \hfill
      \begin{subfigure}{0.31\linewidth}
    \includegraphics[width=\linewidth]{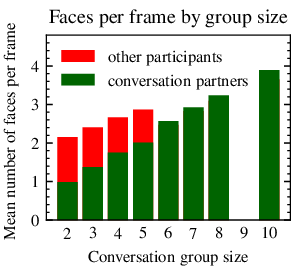}
  \end{subfigure}
  \caption{Descriptive statistics of the dataset.}
  \label{fig:stats}
\end{figure*}

To investigate communication context with egocentric video we collected a unique dataset of people conversing in various simultaneous conversations. The dataset was collected in an experiment of 6 sessions where a total of 48 subjects participated.
%(22F, 1X, 25M; age mean = 25.5, std = 4 years; 3 native English speakers, 45 proficient second language speakers; no visual / auditory / speech / psychiatric conditions / impairments that may affect conversations)
These experiments were approved by the Science-Ethics Committee for the Capital Region of Denmark (reference H-16036391) and all participants provided informed consent to participate.

Per recording session, a group of 6-10 individuals was seated around a rectangular table where they were instructed to engage in conversations in defined subgroups. For each conversation a conversation starter and group assignment were provided by the experimenter. Seating arrangements and conversation groups were pseudo-randomized and changed multiple times per recording session. The conversations were conducted in 1-5 groups of 2-10 people each. Group sizes were chosen to roughly simulate everyday situations such as in a canteen \cite{Dunbar1995SizeGroups, Henzi2007LookCliques}, and also included conversations in a bigger group consisting of all present participants. 50\% of conversations in smaller groups featured some spatial overlap of conversation groups to emphasize scenarios that may be less common but are particularly challenging. Additionally, 50\% of conversations featured multi-talker background noise that was played via 8 loudspeakers placed in a circle around the table (55dBA at the center). Each participant took part in 20 ~5 min conversations, which were balanced with regard to group sizes, background noise and spatial overlap.

Egocentric video data (1080p in color) was collected for each participant during the conversations using Tobii Pro glasses 2 and 3 (Tobii AB, Sweden) or Zetronix Z-shades (Zetronix Corp., US). Figure \ref{fig:frame} shows an example frame of the dataset. The video data was reviewed and cut to only include the actual conversations in the assigned groups. Per frame of the dataset, face bounding boxes were determined using YuNet \cite{Wu2023YuNet:Detector} and were matched to participants using SFace \cite{Zhong2021SFace:Recognition}. Temporal and spatial consistency of the detections was ensured by grouping them into tracklets based on the optical flow between frames \cite{Shi1994GoodTrack, Lucas1981AnVision} and eliminating short tracklets with a low face recognition match.

The resulting dataset consists of 68.9 hours (6.2 million frames) of egocentric video segmented into 877 conversation clips of 4.7 min. on average. The average conversation group size is 4.1 with groups of 2, 3, 4/5 (groups of 5 only occurred with 10 people around the table and are therefore presented with the groups of 4)
and 6 or more individuals making up 25\%, 30\%, 24\% and 20\% of the data respectively resulting in a diverse dataset of conversation scenarios (Figure \ref{fig:stats}). On average there are 2.6 faces in each video frame of. By design the assigned conversation partners per clip are known and make up 66\% of all detected faces.

Additionally to the egocentric video, calibrated eye tracking data was obtained for 74\% (52 hours) of the clips. This data can be used to determine where the camera wearer is looking and may facilitate an evaluation of gaze estimation algorithms on our dataset. With this, the data can be also used to employ and evaluate LAM algorithms from the Ego4D challenges on the data \cite{Grauman2022Ego4D:Video}.

Furthermore, we recorded audio with a 32-channel spatial microphone (mh acoustics LLC, US) placed in the middle of the table. This audio data can be used to evaluate the potential of a spatial beamforming system to separate out the speech audio of conversation partners.

%While egocentric video was collected from all participants at the same time and the clips were aligned to the frame accuracy (+-40ms), the cameras were not genlocked and the audio is not aligned with a higher precision. For the task of identifying conversation context, each video is considered independently. 

The entire dataset is split into a training set, a validation set, a matched test set and an unseen test set, such that all egocentric clips showing the same conversation from different perspectives are always part of the same split. The unseen test set only contains data from one session not used in the other splits. This session was the only session with 8 individuals, which allows to evaluate how models could generalize to new people and slightly different seating arrangements. Overall the training, validation, matched test and unseen test sets account for 40\%, 20\%, 20\% and 19\% of the frames respectively.

For our dataset we formulate the task as: Given egocentric video and face bounding boxes, identify who is part of the camera wearers conversation group. For the current work, we perform this as a binary classification per face per frame and evaluate using the average precision (AP).

\section{Baseline Results}
\label{sec:baseline}

Here, we present our initial baseline classification scores based on simple features without access to temporal context. The baseline results utilize simple heuristics as decision criteria Given a detected face per video frame, we classify the face as a conversation partner based on: center distance (how close is the face to the center of the frame), face bounding box size (how close is the face to the camera wearer), confidence score of the face detection \cite{Wu2023YuNet:Detector}, and similarity score of face recognition \cite{Zhong2021SFace:Recognition}). 
An AP of 0.65 can be achieved by classifying every face as a conversation partner. As shown in Table \ref{tab:results}, all simple decision criteria achieve an AP higher than .7, however the performance of each criterion varies greatly by the subset of the data it is evaluated with. This is a feature of the dataset's diversity of contexts, making identification of conversation partners across contexts and group sizes non-trivial.
\begin{table}[t]
  \centering
  \begin{tabular}{@{}lccccc@{}}
    \toprule
    Test set & \multicolumn{4}{c}{matched} & unseen \\
    Group size & any & 2 & 3 & 4/5 & any \\
    \midrule
    Everyone & 0.65 & 0.41 & 0.52 & 0.65 & 0.64 \\
    Center distance & 0.73 & 0.78 & 0.54 & 0.69 & 0.77 \\
    Face box size & 0.72 & 0.59 & 0.68 & 0.83 & 0.66 \\
    Face det. score & 0.72 & 0.58 & 0.59 & 0.68 & 0.70 \\
    Face rec. match & 0.77 & 0.70 & 0.68 & 0.79 & 0.78 \\
    \bottomrule
  \end{tabular}
  \caption{Average precision (AP) on the task of classifying detected faces as communication partners for different classification criteria and different parts of the test sets. The criterion "Everyone" describes the AP achievable when assuming everyone is part of the same conversation group.
  We do not show the AP for groups of 6+ individuals since in this case everyone was a communication partner. These cases are included in the "any" column.
  }
  \label{tab:results}
\end{table}
\section{Conclusion}
\label{sec:conclusion}

Our work introduces the novel computer vision problem of identifying conversation partners from egocentric video and describes a dataset with baseline measures.

%The initial results demonstrate that even naive methods, which disregard temporal context, yield substantial insights.
%For example, we use the distance of a face to the center as a decision criterion. The rational being that conversation partners are more likely close to the center, which is often used to motivate spatial beamformers in hearing aids that amplify sounds from the front. The baseline results indicate that this may work well if there is only one partner, in groups of 2, but worse in settings with more partners.
Baseline measures vary significantly by the data subset indicating that the occurrence of conversation partners in egocentric video is highly dependent on the situation and group size. The unseen test set is the only part of the dataset with eight people around the table and accordingly slightly different group arrangements and distributions of group sizes compared to the rest of the dataset which is reflected in the different AP scores compared to the matched test set. This illustrates the challenges in generalizing across different conversation scenarios.

%To be clear, we do not assume that people exclusively attend to others in their conversation group nor that in real-world situations people can be strictly attributed to specific conversation groups.
Future work may employ existing approaches, such as predictions of gaze \cite{kellnhofer2019gaze360}, TTM, LAM \cite{Grauman2022Ego4D:Video} or SAAL \cite{Ryan2023EgocentricConversations} on our data. This could establish a direct comparison of the merits of these approaches. However, to identify the conversation partners, perfect knowledge of these measures at every intances may not be required. Since communication groups remains stable for a longer time, we believe that the addition of long temporal context to our models should significantly improve performance.

This work lays the foundation for future investigations, contributing to a deeper understanding of conversations in complex environments. Such analyses could inform decisions about spatial beamforming or selective speech separation in hearing aids. %Our findings indicate that the assumption that communication targets are mainly distributed directly in front of the listener may only be valid in relatively simple communication situations. Moreover, our data could be used to investigate the potential and challenges in the development of intelligent hearing devices that leverage gaze or video data to steer acoustic processing.

{
    \small
    \bibliographystyle{ieeenat_fullname}
    \bibliography{references}
}


\end{document}